\documentclass[sigconf]{acmart}

\AtBeginDocument{%
  }

\copyrightyear{2026}
\acmYear{2026}
\setcopyright{cc}
\setcctype{by-nc-nd}
\acmConference[ETRA '26]{2026 Symposium on Eye Tracking Research and Applications}{June 01--04, 2026}{Marrakesh, Morocco}
\acmBooktitle{2026 Symposium on Eye Tracking Research and Applications (ETRA '26), June 01--04, 2026, Marrakesh, Morocco}
\acmDOI{10.1145/3797246.3803050}
\acmISBN{979-8-4007-2519-7/2026/06}

\usepackage{algorithmic}
\usepackage{graphicx}
\usepackage{textcomp}
\usepackage{xcolor}
\usepackage{booktabs}
\usepackage{multirow}
\usepackage{array}
\usepackage{url}
\usepackage{hyperref}
\usepackage{algorithm}
\usepackage{balance}
\usepackage{xspace}
\usepackage{natbib}

\begin{document}

% \title{Factor-Based Uncertainty Estimation for Gaze Prediction}
\title{Factor-Informed Uncertainty Distillation for Gaze Estimation}

% Anonymous submission hygiene: omit author blocks entirely.
% \renewcommand{\shortauthors}{Anonymous}

\author{Mohammadreza Jamalifard}
\authornote{Mohammadreza Jamalifard and Yaxiong Lei contributed equally to this work.}
\email{m.jamalifard@essex.ac.uk}
\orcid{0009-0008-4969-1398}
\affiliation{%
  \institution{University of Essex}
  \city{Colchester}
  \country{UK}
}

\author{Yaxiong Lei}
\authornotemark[1]
\authornote{Corresponding authors: Yaxiong Lei and Javier Andreu-Perez.}
\email{yaxiong.lei@essex.ac.uk}
\orcid{0000-0002-0697-7942}
\affiliation{%
  \institution{University of Essex}
  \city{Colchester}
  \country{UK}
}

\author{Javier Fumanal Idocin}
\email{j.fumanal-idocin@essex.ac.uk}
\orcid{0000-0002-0644-1355}
\affiliation{%
  \institution{University of Essex}
  \city{Colchester}
  \country{UK}
}

\author{Parastoo Azizinezhad}
\email{p.azizinezhad@essex.ac.uk}
\orcid{0009-0008-2537-5059}
\affiliation{%
  \institution{University of Essex}
  \city{Colchester}
  \country{UK}
}

\author{Tom Foulsham}
\email{foulsham@essex.ac.uk}
\orcid{0000-0002-8444-7269}
\affiliation{%
  \institution{University of Essex}
  \city{Colchester}
  \country{UK}
}

\author{Javier Andreu-Perez}
\authornotemark[2]
\email{j.andreu-perez@essex.ac.uk}
\orcid{0000-0002-7421-4808}
\affiliation{%
  \institution{University of Essex}
  \city{Colchester}
  \country{UK}
}

\begin{abstract}
Deep gaze estimation works well in controlled capture but degrades in unconstrained settings, where systems must reject unreliable predictions. Single-pass uncertainty (e.g., heteroscedastic regression) infers uncertainty from pixels without explicit input-validity cues, while sampling based methods are often too costly for real time use. We propose Factor-Informed Uncertainty Distillation (FIUD), a teacher-student framework that aligns uncertainty with interpretable image-quality failure modes. A gradient-boosting teacher predicts expected gaze error from factors such as illumination, sharpness, eye visibility and symmetry; a neural student distills these signals via curriculum learning and ranking supervision into a lightweight single-pass uncertainty head. Across ETH-XGaze, Gaze360, and MPIIFaceGaze ($>$300k samples), FIUD improves uncertainty, error rank correlation and selective prediction versus deterministic and sampling-based baselines, with the largest gains in unconstrained settings.

% Deep gaze estimation works well in controlled capture but degrades in unconstrained settings, where systems must reject unreliable predictions. Single-pass uncertainty (e.g., heteroscedastic regression) infers uncertainty from pixels without explicit input-validity cues, while sampling based methods are often too costly for real time use.
% We propose Factor-Informed Uncertainty Distillation (FIUD), a teacher-student framework that aligns uncertainty with interpretable image-quality failure modes. A gradient-boosting teacher predicts expected gaze error from factors such as illumination, sharpness, and eye visibility; a neural student distills these signals via curriculum learning and ranking supervision into a lightweight single-pass uncertainty head.
% Across ETH-XGaze, Gaze360, and MPIIFaceGaze ($>$300k samples), FIUD improves uncertainty, error rank correlation and selective prediction versus deterministic and sampling-based baselines, with the largest gains in unconstrained settings.

\end{abstract}

% \ccsdesc[300]{Human-centered computing~Interaction techniques}


\maketitle

\section{Introduction}
Gaze estimation predicts line of sight from facial images and supports applications in human-computer interaction across desktop, mobile, and headsets~\cite{villanueva2008novel, lei2023end, cheng2024appearance, lei2026people, lei2026gazecode, lei2026tinygaze, lei2026gazesync}, driver monitoring~\cite{vicente2015driver}, augmented reality~\cite{patney2016towards}, and privacy-aware sensing~\cite{christina2020role, he2025identity, lei2023protecting,leiprotecting}. Despite strong accuracy in controlled capture, performance degrades in unconstrained settings~\cite{zhong2024uncertainty, lei2025quantifying}, where systems must avoid confidently acting on erroneous gaze. In safety- and interaction-critical scenarios, the ability to \emph{abstain} on unreliable inputs is therefore essential.

Uncertainty quantification (UQ) is a natural mechanism for rejection. However, sampling-based methods such as Deep Ensembles~\cite{lakshminarayanan2017simple} and MC Dropout~\cite{gal2016dropout} require multiple forward passes and are often impractical for real-time deployment. Deterministic single-pass approaches are efficient~\cite{mukhoti2023deep}, but heteroscedastic regression~\cite{kendall2017uncertainties} must infer uncertainty implicitly from pixels, without explicit supervision about common invalid-input failure modes (e.g., low illumination, blur, occlusion, extreme pose). In many gaze interfaces, uncertainty is primarily used as a \emph{ranking signal} for selective prediction (accept/reject), so aligning uncertainty with these failure modes is more useful than estimating a perfectly calibrated variance.

We propose \emph{Factor-Informed Uncertainty Distillation (FIUD)}, a teacher--student framework that trains uncertainty to reflect interpretable image-quality causes of error while retaining single-pass inference. A gradient-boosting teacher predicts expected gaze error from quality factors (illumination, sharpness, eye visibility/symmetry), and a neural student distills this signal using curriculum learning and ranking supervision into a lightweight uncertainty head. The factors are used only during training; at inference, FIUD requires no factor extraction and matches deterministic latency.

Our contributions are: \textbf{(1)} a factor-based teacher that provides an interpretable quality-to-error supervision signal for gaze reliability; \textbf{(2)} a single-pass student trained via distillation and ranking to optimize uncertainty ordering for selective prediction; and \textbf{(3)} a deployment-oriented evaluation on ETH-XGaze, Gaze360, and MPIIFaceGaze showing improved uncertainty error correlation and sparsification performance over deterministic and sampling-based baselines.
%The remainder of the paper is organized as follows: Sec.~2 reviews related work; Sec.~3 describes FIUD; Sec.~4 details experimental setup; Sec.~5 presents results; Sec.~6 discusses findings; Sec.~7 lists limitations; and Sec.~8 concludes.

\section{Related Work}

\subsection{Deep Learning for Gaze Estimation}
Appearance-based gaze estimation has evolved from geometric methods relying on explicit feature detection to end-to-end deep learning models~\cite{zhang2015appearance, cheng2024appearance, lei2023end}. Modern architectures leverage convolutional neural networks and vision transformers~\cite{cheng2024appearance, lei2023end} trained on large-scale datasets, e.g. MPIIGaze~\cite{zhang2017mpiigaze}, GazeCapture~\cite{krafka2016eye}, ETH-XGaze~\cite{zhang2020ethxgaze}, and Gaze360~\cite{kellnhofer2019gaze360}. While extensive research focuses on improving accuracy via architecture~\cite{cheng20253d, liu2025gaze, qin2025unigaze}, domain adaptation~\cite{zheng2025enhancing, cai2023source, qin2025domain, wang2025ptgaze}, and continual learning~\cite{lei2025mac, sugano2008incremental}
% and knowledge distillation~\cite{guo2019generalized, chen2025large}
, fewer studies explicitly address the challenge of quantifying reliability during deployment~\cite{zhong2024uncertainty, lei2025quantifying, hu2025unmode}. Our work targets this gap, prioritizing actionable uncertainty for real-world interaction.

\subsection{Uncertainty Estimation in Deep Learning}
% Standard uncertainty quantification (UQ) approaches often trade off efficiency and reliability. Bayesian neural networks and sampling-based approximations like Deep Ensembles~\cite{lakshminarayanan2017simple} or MC Dropout~\cite{gal2016dropout} offer robust estimates but require multiple forward passes, incurring prohibitive latency for real-time gaze interfaces. Conversely, deterministic methods such as heteroscedastic regression~\cite{kendall2017uncertainties} enable single-pass inference by predicting variance via negative log-likelihood (Eq.~\ref{eq:nll}). However, these methods must implicitly learn "input validity" solely from pixel noise without explicit supervision, leading to overconfident predictions on out-of-distribution samples. While knowledge distillation has been applied to UQ~\cite{malinin2020ensemble}, prior works typically distill an ensemble's statistical variance (a black-box signal) into a single network. In contrast, our approach distills \emph{interpretable, cause-and-effect knowledge} regarding image quality, explicitly grounding uncertainty in physical failure modes.

Standard approaches to uncertainty quantification (UQ) often trade off between computational efficiency and reliability. Bayesian neural networks and sampling-based approximations like Deep Ensembles~\cite{lakshminarayanan2017simple} or MC Dropout~\cite{gal2016dropout} offer robust uncertainty estimates but require multiple forward passes, incurring latency costs often prohibitive for real-time gaze interfaces. Conversely, deterministic methods such as heteroscedastic regression~\cite{kendall2017uncertainties} enable single-pass inference by predicting variance via negative log-likelihood (Eq.~\ref{eq:nll}). However, these methods must implicitly learn the concept of "input validity" solely from pixel noise without explicit supervision, which can lead to overconfident predictions on out-of-distribution samples. While knowledge distillation and explainable symbolic AI tools have been explored~\cite{malinin2020ensemble,fumanal2024exfuzzy}, prior works typically distill the statistical variance of an ensemble (a black-box signal) into a single network. In contrast, our approach distills \emph{interpretable, cause-and-effect knowledge} regarding image and feature quality, grounding uncertainty in physical failure modes.
\begin{equation}
\mathcal{L}_{\text{NLL}} = \frac{1}{2\sigma^2}(y - \mu)^2 + \frac{1}{2}\log(\sigma^2)
\label{eq:nll}
\end{equation}

% \subsection{Uncertainty Estimation in Deep Learning}
% Bayesian neural networks are often computationally expensive in practice~\cite{blundell2015weight}. Deep Ensembles~\cite{lakshminarayanan2017simple} and MC Dropout~\cite{gal2016dropout} provide practical approximations but typically require multiple forward passes. Heteroscedastic regression predicts mean and variance directly using the negative log-likelihood loss:
% \begin{equation}
% \mathcal{L}_{\text{NLL}} = \frac{1}{2\sigma^2}(y - \mu)^2 + \frac{1}{2}\log(\sigma^2)
% \label{eq:nll}
% \end{equation}
% This enables single-pass inference, but uncertainty must be learned from pixels without explicit supervision about input validity. Furthermore, recent work extends distillation to uncertainty estimation~\cite{malinin2020ensemble}. Essentially, knowledge distillation transfers knowledge from teacher models to compact students~\cite{hinton2015distilling}, often distilling ensemble outputs into single networks.

\subsection{Image Quality Assessment for Gaze Estimation}
Image quality assessment is a standardized component of iris recognition pipelines~\cite{grother2008biometrics} but remains underutilized in gaze estimation. Quality measures have also been used as practical decision signals in face biometrics, e.g., for presentation-attack (anti-spoofing) detection using IQA cues~\cite{fourati2020iqaantispoof}. While general-purpose no-reference metrics like BRISQUE~\cite{mittal2012no} and NIQE~\cite{mittal2013making} capture perceptual distortions, they often fail to reflect the specific geometric ambiguities that degrade gaze accuracy (e.g., a sharp image with obscured pupils). Prior gaze research has acknowledged the performance drop caused by low-quality inputs~\cite{cheng2024appearance, lei2023end} and explored mitigation strategies including specialized network architecture variations~\cite{ansari2023person, cheng2022gaze, ververas20243dgazenet} and calibration techniques~\cite{lei2025mac, sugano2015appearance}. However, these works largely treat quality variations as a covariate to be robust against. The explicit mapping between interpretable gaze-specific quality factors, e.g. eye openness or local contrast, and gaze reliability remains underexplored. This work addresses that gap by leveraging such factors not as inputs, but as supervisory signals to shape the uncertainty space of deep gaze networks.

% \subsection{Image Quality Assessment for Gaze Estimation}
% Image quality assessment is a standardized component of iris recognition pipelines~\cite{iso19795} but remains underutilized in gaze estimation. While general-purpose no-reference metrics like BRISQUE~\cite{mittal2012no} and NIQE~\cite{mittal2013making} capture perceptual distortions, they often fail to reflect the specific geometric ambiguities that degrade gaze accuracy. Prior gaze research has acknowledged the performance drop caused by low-quality inputs~\cite{cheng2024appearance, lei2023end} and explored mitigation strategies including specialized network architecture variations~\cite{ansari2023person, cheng2022gaze, ververas20243dgazenet} and calibration techniques~\cite{lei2025mac, sugano2015appearance}. However, the explicit mapping between interpretable gaze-specific quality factors—such as eye openness or local contrast—and gaze reliability remains underexplored. This work addresses that gap by leveraging such factors to quantify uncertainty.

% Image-quality assessment is widely studied in iris recognition~\cite{iso19795}. General-purpose no-reference metrics such as BRISQUE~\cite{mittal2012no} and NIQE~\cite{mittal2013making} capture perceptual quality but are not tailored to gaze estimation. Quality-aware training has been explored for face recognition~\cite{shi2019probabilistic, lemley2019convolutional}, yet gaze-specific quality factors remain underexplored. Our work aims to connect interpretable quality cues to gaze reliability.

\section{Methodology}

\subsection{Overview}
\begin{figure*}[ht]
\centering
\includegraphics[width=0.95\linewidth]{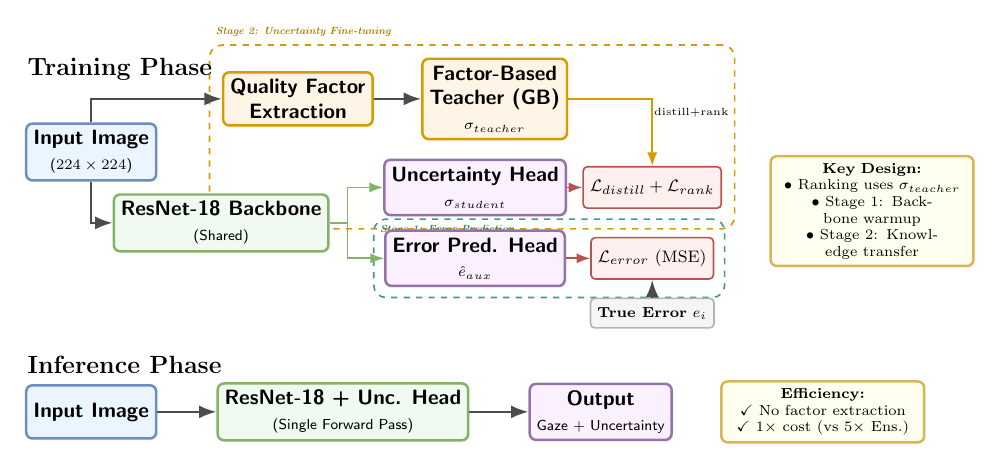}
\caption{FIUD framework. \textbf{Training:} quality factors are extracted via MediaPipe and used to train a gradient-boosting teacher. The student is trained with curriculum learning: Stage 1 learns error-predictive representations; Stage 2 distills teacher signals into an uncertainty head. \textbf{Inference:} only the student backbone and uncertainty head are used (no factor extraction).}
\label{fig:framework}
\end{figure*}

FIUD operates in two phases (Fig.~\ref{fig:framework}). During training, we extract quality factors and train an interpretable factor-based teacher to predict expected gaze error from these factors. We then train a neural student uncertainty head using curriculum learning and teacher-derived supervision. During inference, only the student is used, providing single-pass uncertainty without computing factors.

\subsection{Quality Factor Extraction}
To define a practical ``quality envelope'' for gaze inputs, we extract a set of interpretable factors across six categories (Table~\ref{tab:factors}). These include \textbf{illumination and sharpness} (intensity statistics, blur), \textbf{geometric and pose} factors (face scale, head orientation), and \textbf{eye/iris features} (eye aspect ratio, symmetry, pupil visibility). These factors are interpretable by design (e.g., low face brightness $\rightarrow$ poor lighting), enabling actionable feedback. MediaPipe extraction succeeds for 93.3\% (ETH-XGaze), 94.8\% (Gaze360), and 97.5\% (MPIIFaceGaze).

% \subsection{Quality Factor Extraction}
% To define a practical ``quality envelope'' for gaze inputs, we extract a set of interpretable factors across six categories (Table~\ref{tab:factors}). \textbf{Illumination and sharpness} quantify image quality via intensity statistics, Laplacian variance, and texture coherence (face clarity). \textbf{Geometric and pose} factors encode face scale and head orientation derived from MediaPipe landmarks when available. \textbf{Eye and iris} features capture visibility cues, including Eye Aspect Ratio (EAR), bilateral symmetry, iris diameter, and pupil visibility (unobstructed iris pixel proportion). These factors are interpretable by design (e.g., low face brightness $\rightarrow$ poor lighting), enabling actionable feedback. MediaPipe extraction succeeds for 93.3\% (ETH-XGaze), 94.8\% (Gaze360), and 97.5\% (MPIIFaceGaze).

% These factors are interpretable by design: for example, high uncertainty associated with low face brightness corresponds to poor lighting, enabling actionable feedback in interactive systems. All these features extracted from MediaPipe, and its detection rates: 93.3\% (ETH-XGaze), 94.8\% (Gaze360), and 97.5\% (MPIIFaceGaze).

% \textbf{Detection success rates.} MediaPipe Face Mesh achieves high landmark detection rates across datasets: 93.3\% (ETH-XGaze), 94.8\% (Gaze360), and 97.5\% (MPIIFaceGaze). Training uses samples where factor extraction succeeds.

\begin{table}[ht]
\centering
\resizebox{0.95\columnwidth}{!}{
\begin{tabular}{ll}
\toprule
\textbf{Category} & \textbf{Factors} \\
\midrule
Illumination      & face brightness, brightness variation \\
Sharpness         & sharpness, sharpness variation, face clarity \\
Geometry          & face area ratio, face depth proxy \\
Head Pose         & head pose pitch, head pose yaw, head pose roll \\
Eye Features      & eye aspect ratio (left), eye aspect ratio (right), eye symmetry \\
Iris/Pupil        & iris diameter (L/R), pupil visibility (L/R) \\
\bottomrule
\end{tabular}}
\caption{Quality factors for gaze uncertainty estimation.}
\label{tab:factors}
\end{table}

\subsection{Factor-Based Teacher}
The teacher is a gradient-boosting regressor trained to predict angular gaze error from quality factors:
\begin{equation}
\sigma_{\text{teacher}} = f_{\text{GB}}(x_1, x_2, \ldots, x_{d})
\end{equation}
where $d$ is the number of available factors after dataset-specific filtering. The teacher requires explicit factor extraction; FIUD distills this quality-aware signal into a student that does not require factors at inference.

\textbf{Role of the teacher.} The teacher is not intended to be a strong standalone predictor of gaze error; rather, it provides a structured, interpretable \emph{quality-to-error} signal that can be distilled into the student. Even a coarse but systematic ordering over samples is valuable for selective prediction, where uncertainty is used primarily as a ranking signal. The student can combine this distilled structure with appearance cues unavailable to the factor-only teacher, yielding uncertainty estimates that better align with observed failures while remaining efficient at inference.

\textbf{Leakage-free evaluation.} The teacher is trained on the training split only. During student training, teacher outputs are used only for training/validation samples. All reported test metrics are computed using the student on the held-out test split, and teacher performance (Spearman $\rho$) is reported on the held-out test split: 0.1434 (ETH-XGaze), 0.4947 (Gaze360), and 0.1705 (MPIIFaceGaze). This indicates that factor-only predictability is strongest in unconstrained in-the-wild data and weaker in cross-subject settings where subject-specific variance contributes to error.

% \textbf{Leakage-free evaluation.} The teacher is trained on the training split only. During student training, teacher outputs are used only for training/validation samples. All reported test metrics are computed using the student on the held-out test split, and teacher performance (Spearman $\rho$) is reported on the held-out test split for transparency: 0.1434 (ETH-XGaze), 0.4947 (Gaze360), and 0.1705 (MPIIFaceGaze).This indicates that factor-only predictability is strongest in unconstrained in-the-wild data and weaker in cross-subject settings where subject-specific variance contributes to error.

\textbf{Teacher-Student Generalization.} While the teacher is limited to samples with successful landmark detection, the student model processes raw pixel intensities directly. This allows the student to potentially learn appearance-based failure cues (e.g., extreme motion blur or total occlusion) that cause landmark detectors to fail, effectively generalizing the "quality-to-error" logic to the most challenging edge cases where factor extraction is impossible\cite{malinin2020ensemble}.

\subsection{Student Architecture}
The student uses a ResNet-18 backbone (ImageNet pretrained) with a gaze head predicting pitch and yaw and an uncertainty head predicting scalar uncertainty. The backbone produces a 512-D feature vector $f_\theta(I)$:
\begin{align}
\hat{g} &= h_{\text{gaze}}(f_\theta(I)) \\
\sigma_{\text{student}} &= \text{Softplus}(h_{\text{unc}}(f_\theta(I)))
\end{align}
The uncertainty head is a 3-layer MLP (512$\rightarrow$256$\rightarrow$128$\rightarrow$1) with batch normalization and dropout ($p=0.3$). For curriculum learning, we add an auxiliary error-prediction head with the same architecture.

\subsection{Training Objective}
We train the uncertainty head with a composite loss:
\begin{equation}
\mathcal{L}_{\text{total}} = \lambda_{\text{distill}} \mathcal{L}_{\text{distill}} + \lambda_{\text{rank}} \mathcal{L}_{\text{rank}}
\end{equation}
Distillation aligns student uncertainty to teacher-predicted error:
\begin{equation}
\mathcal{L}_{\text{distill}} = \frac{1}{N} \sum_{i=1}^{N} (\sigma_{\text{student}}^{(i)} - \sigma_{\text{teacher}}^{(i)})^2
\end{equation}
Ranking optimizes uncertainty ordering using the teacher as an ordering signal:

\begin{equation}
\mathcal{L}_{\text{rank}} = \frac{1}{|P|} \sum_{(i,j) \in P} \max(0, m - s_{ij}(\sigma_i - \sigma_j))
\label{eq:7}
\end{equation}
where $s_{ij} = \text{sign}(\sigma_{\text{teacher}}^i - \sigma_{\text{teacher}}^j)$ and $m$ is a margin. We set $\lambda_{\text{distill}} = 0.5$, $\lambda_{\text{rank}} = 0.3$, and $m = 0.1$ based on validation performance.

\subsection{Training Strategy}
We use curriculum learning to stabilize training and encourage the backbone to learn error-predictive representations.

\textit{Stage 1: Error-prediction warm-up (20 epochs).} We train an auxiliary error head with ground-truth angular error:
\begin{equation}
\mathcal{L}_{\text{stage1}} = \text{MSE}(\hat{e}_{\text{aux}}, e_{\text{true}})
\end{equation}
This is a training-only auxiliary objective (analogous to common auxiliary losses in gaze estimation) that encourages error-aware representations early in training; it does not introduce any additional inference-time supervision.

\textit{Stage 2: Uncertainty distillation (50 epochs).} We train the uncertainty head using teacher-derived supervision while retaining the auxiliary task:
\begin{equation}
\mathcal{L}_{\text{stage2}} = \lambda_{\text{distill}} \mathcal{L}_{\text{distill}} + \lambda_{\text{rank}} \mathcal{L}_{\text{rank}} + \lambda_{\text{aux}} \mathcal{L}_{\text{error}}
\end{equation}

\textbf{Clarification of supervision.} Stage 1 uses supervised angular error during training to encourage error-aware representations; this signal is derived from the same ground-truth gaze labels used in standard supervised gaze training. Stage 2 uses the factor-based teacher for distillation and ranking supervision. At inference time, only the student backbone and uncertainty head are used; no quality factors or teacher models are required.

\subsection{Design Rationale}
FIUD combines (i) a factor-based teacher and (ii) a student trained with both regression-to-teacher and ranking supervision. We use an MSE distillation loss to transfer the teacher's coarse \emph{quality-to-error} signal into the student uncertainty head, while the margin-based ranking loss directly optimizes the \emph{ordering} required for selective prediction. Curriculum learning stabilizes optimization by first encouraging the backbone to represent error-predictive features before introducing teacher-guided ranking constraints. We view these design choices as complementary: the teacher provides interpretable structure, the student can exploit richer appearance cues, and the ranking objective aligns training with deployment-time accept/reject decisions. A full ablation of these components is left to future work.

\section{Experimental Setup}

\subsection{Datasets}
We evaluate on three benchmarks representing controlled, in-the-wild, and personal computing scenarios (Table~\ref{tab:datasets}). \textbf{ETH-XGaze}~\cite{zhang2020ethxgaze} provides controlled, high-resolution capture. \textbf{Gaze360}~\cite{kellnhofer2019gaze360} includes unconstrained indoor/outdoor settings with extreme head pose. \textbf{MPIIFaceGaze}~\cite{zhang2017mpiigaze} contains webcam images; we use leave-one-person-out cross-validation to evaluate generalization to unseen subjects.

\textbf{Evaluation subset and filtering.} To ensure a fair comparison, all methods are trained and evaluated on the subset of samples where MediaPipe landmark detection succeeds. While this excludes cases of total detection failure (e.g., complete occlusion), it isolates the challenge of \textit{appearance-based reliability estimation}: distinguishing between high-quality inputs and those that are detectable but degraded by blur, noise, or extreme pose.

% \textbf{Evaluation subset and filtering.} Because FIUD's teacher requires successful factor extraction, we train and evaluate \emph{all} methods on a consistent subset where MediaPipe landmark detection succeeds. This ensures a like-for-like comparison across uncertainty methods under the same input availability assumptions (i.e., a face/landmark detector operating upstream). We report dataset statistics and metrics on the subset.
% this MediaPipe-success subset.

\textbf{Mean error reporting (within-protocol).} We report mean angular error as computed/loaded by our pipeline from the provided inference files after MediaPipe filtering. Absolute mean errors on Gaze360 may differ from some commonly cited values due to differences in preprocessing and file conventions; our comparisons are therefore intended to be \emph{within-protocol} across methods under the same data source and filtering.

\begin{table}[ht]
\centering
% \caption{Dataset characteristics (pipeline statistics after  filtering; i.e., the MediaPipe-success subset used for training and evaluation).}
\resizebox{0.95\columnwidth}{!}{
\begin{tabular}{@{}lcccc@{}}
\toprule
\textbf{Dataset} & \textbf{Samples} & \textbf{Type} & \textbf{Mean Err.} & \textbf{Protocol} \\
\midrule
ETH-XGaze & 140,598 & Controlled & 5.08$^\circ$ & Subject-disjoint \\
Gaze360 & 106,467 & In-the-wild & 32.44$^\circ$ & Official split \\
MPIIFaceGaze & 36,737 & Webcam & 16.36$^\circ$ & LOOCV \\
\bottomrule
\end{tabular}}
\caption{Dataset characteristics (pipeline statistics after  filtering).}
\label{tab:datasets}
\end{table}

\subsection{Baselines}
We compare FIUD against common UQ categories. \textbf{Deep Ensembles}~\cite{lakshminarayanan2017simple} estimate uncertainty from prediction variability across multiple independently trained models (multiple forward passes at inference). \textbf{MC Dropout}~\cite{gal2016dropout} estimates uncertainty by performing stochastic forward passes with dropout enabled at inference. \textbf{Heteroscedastic Regression (NLL)}~\cite{kendall2017uncertainties} predicts mean and variance in a single pass using Eq.~\ref{eq:nll}.

\subsection{Evaluation Metrics}
We use complementary metrics that reflect deployment needs. \textit{Spearman rank correlation ($\rho$)} measures whether higher predicted uncertainty aligns with higher observed gaze error. \textit{Area Under Sparsification Error (AUSE)}~\cite{ilg2018uncertainty} measures selective prediction utility: how quickly error decreases when rejecting high-uncertainty samples compared to an oracle. We evaluate uncertainty primarily as a \emph{ranking} signal for rejection. From an HCI perspective, the \emph{relative ranking} of predictions is often more vital than absolute probabilistic calibration \cite{mukhoti2023deep}. A gaze-based system primarily needs to know which $N\%$ of frames to ignore to maintain a fluid interaction. Consequently, we prioritize Spearman correlation and AUSE, as they directly measure the model's ability to prioritize high-quality inputs for the downstream application.

\subsection{Implementation Details}
All models use ResNet-18 backbones pretrained on ImageNet and fine-tuned with 224$\times$224 input. The uncertainty head is trained with AdamW (LR $10^{-4}$, weight decay $10^{-5}$), batch size 128. Curriculum learning uses a higher LR during Stage 1 and a lower backbone LR during Stage 2. We ran three random seeds (42/123/456). Due to a deterministic evaluation pipeline and reporting at 4 decimal places, the reported test metrics were effectively identical across seeds; we therefore report a single value per method.

\section{Results and Analysis}

\subsection{UQ Method Comparison}
Table~\ref{tab:uq_comparison} compares uncertainty quality across methods. FIUD improves the alignment between predicted uncertainty and observed error (Spearman $\rho$) and improves selective prediction utility (AUSE) in aggregate, with the largest gains in the unconstrained Gaze360 setting. These results support the value of incorporating interpretable quality cues during training for deployment-oriented uncertainty estimation. The performance disparity between controlled (ETH-XGaze) and in-the-wild (Gaze360) datasets highlights a key characteristic of FIUD. In regulated, high-resolution settings, gaze error is often driven by stochastic sensor noise or subtle subject-specific anatomical traits which are inherently difficult to capture via general quality factors \cite{lei2025quantifying}. Conversely, in unconstrained settings, failure is dominated by systematic environmental factors (illumination, pose, blur). FIUD specifically targets these actionable failures, leading to the substantial gains observed in Gaze360 ($\rho=0.4980$), where reliability quantification is critical for system safety \cite{vicente2015driver}.

\begin{table*}[ht]
\centering
\begin{minipage}[t]{0.45\linewidth}
    \centering
    \resizebox{0.9\columnwidth}{!}{
    \begin{tabular}{@{}lcccc@{}}
    \toprule
    \textbf{Dataset} & \textbf{Ens.} & \textbf{MC} & \textbf{NLL} & \textbf{FIUD} \\
    \midrule
    ETH-XGaze & -0.0223 & -0.0581 & 0.1188 & \textbf{0.1319} \\
    Gaze360 & 0.2485 & 0.2581 & 0.3242 & \textbf{0.4980} \\
    MPIIFaceGaze & 0.0663 & 0.0496 & -0.0125 & \textbf{0.1682} \\
    \midrule
    \textit{Average} & 0.0975 & 0.0832 & 0.1435 & \textbf{0.2660} \\
    \bottomrule
    \end{tabular}
    }
    \caption{Uncertainty quality (Spearman $\rho$ on the MediaPipe-success test subset; higher is better).}
    \label{tab:uq_comparison}
\end{minipage}
\hfill
\begin{minipage}[t]{0.45\linewidth}
    \centering
    % \caption{Selective prediction (AUSE $\downarrow$ on the MediaPipe-success test subset; lower is better).}
    \resizebox{0.9\columnwidth}{!}{
    \begin{tabular}{@{}lcccc@{}}
    \toprule
    \textbf{Dataset} & \textbf{Ens.} & \textbf{MC} & \textbf{NLL} & \textbf{FIUD} \\
    \midrule
    ETH-XGaze & 2.4411 & 2.7217 & \textbf{2.0258} & 2.0915 \\
    Gaze360 & 10.4180 & 10.3665 & 9.0865 & \textbf{7.3911} \\
    MPIIFaceGaze & 5.6067 & 5.9703 & 6.2210 & \textbf{5.2106} \\
    \midrule
    \textit{Average} & 6.1553 & 6.3528 & 5.7778 & \textbf{4.8977} \\
    \bottomrule
    \end{tabular}
    }
    \caption{Selective prediction performance (AUSE in degrees; lower is better).}
    \label{tab:ause}
\end{minipage}
\end{table*}

\subsection{Interpretability and System Usefulness}
FIUD distills a factor-to-error relationship into a single-pass uncertainty head, enabling actionable behavior: high uncertainty can trigger abstention, re-capture, or user feedback (e.g., improve lighting or reduce occlusion). Interpretability comes from the \emph{supervisory signal}: the teacher links human-interpretable quality factors to expected error, and distillation aligns the student's uncertainty ranking with these failure modes. At inference, the student requires no factor extraction or factor outputs, supporting real-time deployment while producing uncertainty correlated with actionable capture conditions.

\subsection{Computational Efficiency}
Table~\ref{tab:efficiency} summarizes inference requirements. FIUD provides single-pass uncertainty estimation at inference with latency comparable to NLL, whereas ensembles and MC Dropout require multiple forward passes. In our runtime benchmarks (batch size 128 on GPU), single-pass methods were $\approx$0.15--0.16\,ms per batch, while ensembles (5$\times$) were $\approx$0.51\,ms and MC Dropout (30$\times$) $\approx$2.5\,ms.

\begin{table}[ht]
\centering
\resizebox{0.95\linewidth}{!}{
\begin{tabular}{@{}lcccc@{}}
\toprule
\textbf{Method} & \textbf{Passes} & \textbf{Latency (ms)} & \textbf{Relative} & \textbf{Interp.} \\
\midrule
Deep Ensembles & 5 & $\sim$0.51 & $\sim$3.2$\times$ & No \\
MC Dropout & 30 & $\sim$2.5 & $\sim$16$\times$ & No \\
NLL & 1 & $\sim$0.15--0.16 & 1.0$\times$ & No \\
\textbf{FIUD (Ours)} & \textbf{1} & $\sim$0.15--0.16 & \textbf{1.0$\times$} & \textbf{Yes} \\
\bottomrule
\end{tabular}
}
\caption{Computational efficiency comparison (measured; latency excludes preprocessing and reflects forward-pass cost under the same runtime conditions).}
\label{tab:efficiency}
\end{table}

\subsection{Calibration Analysis}
\begin{figure}[ht]
\centering
\includegraphics[width=0.95\columnwidth]{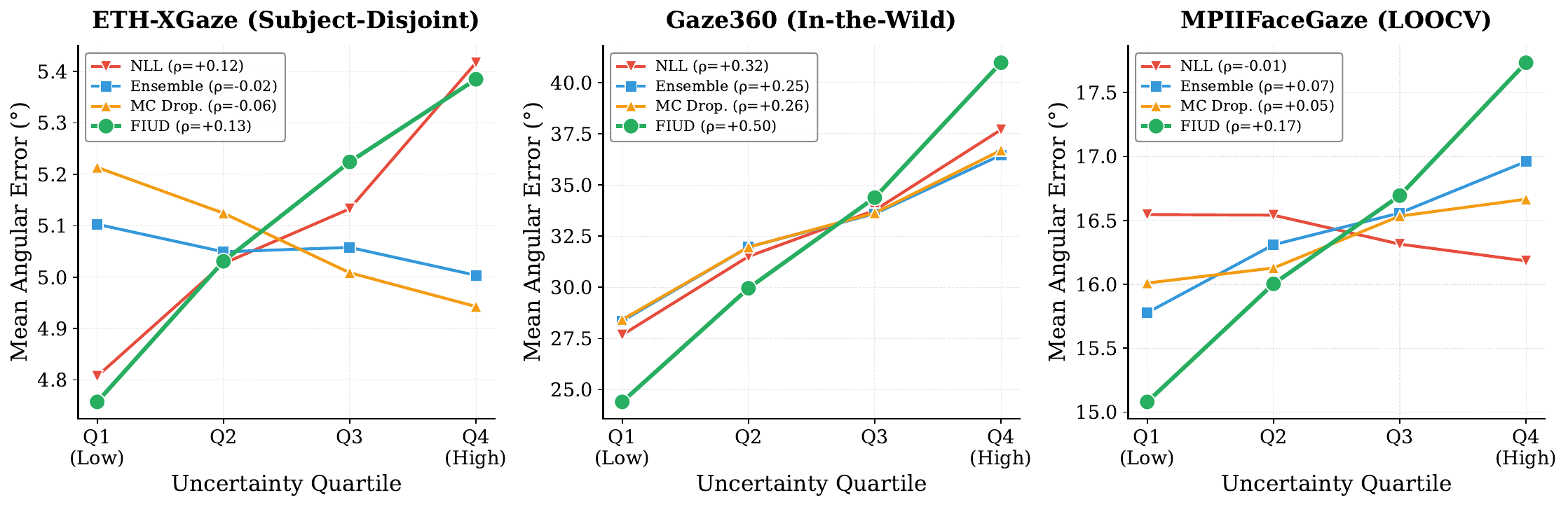}
% \caption{Reliability diagrams across methods and datasets. Trends are most informative as a ranking signal for selective prediction; quantitative utility is reported via AUSE.}
\caption{Reliability diagrams across methods and datasets (ranking-oriented; quantitative utility via AUSE).}
\label{fig:reliability}
\end{figure}

We visualize reliability by partitioning samples into bins by predicted uncertainty and plotting observed error (Fig.~\ref{fig:reliability}). A useful uncertainty signal for selective prediction should assign higher uncertainty to higher-error predictions. We treat calibration primarily through the lens of selective prediction utility (AUSE), since many interactive systems use uncertainty as a ranking signal for abstention.

\section{Discussion}
Our results highlight that explicitly modeling the relationship between image quality and gaze error creates a stronger inductive bias for uncertainty estimation than purely pixel-derived methods. This advantage is most pronounced in unconstrained settings like Gaze360, where error is driven by systematic environmental factors, such as poor illumination or extreme pose---that the teacher successfully identifies and transfers to the student. In contrast, pixel-only methods like NLL must infer these relationships implicitly, often struggling to distinguish between actionable failure modes and irreducible sensor noise. By internalizing the \emph{visual signatures} of these high-error factors through ranking supervision, the student learns to reliably flag low-quality inputs without requiring the expensive factor extraction pipeline at inference.

From an operational perspective, FIUD resolves the tension between reliability and latency by decoupling the computational cost of interpretability from its runtime utility. Because the student produces uncertainty in a single forward pass, it fits strict real-time budgets ($<$1\,ms) while offering the ranking fidelity required for safety-critical "Hard Rejection" policies, such as gating unreliable frames in driver monitoring. This efficiency also enables "Graceful Degradation" in interactive systems: rather than failing silently, a UI can use the uncertainty signal to dynamically expand button hit-boxes~\cite{patney2016towards}, preserving usability during transient sensing failures.

While validated on gaze estimation, this framework offers a generalizable blueprint for perception tasks where failure modes are semantically definable but costly to compute. The ability to distill "quality-to-error" logic into a lightweight head suggests that future work could extend this approach to other domains, such as head pose or emotion recognition.

\section{Limitations}
FIUD is bounded by the teacher's ability to predict error from the selected quality factors. In cross-subject settings (e.g., MPIIFaceGaze), subject-specific characteristics contribute substantially to error variance and may be weakly captured by the defined factor vocabulary. Training depends on successful landmark detection; we train and evaluate on the subset of samples where factor extraction succeeds to ensure consistent comparisons across methods, which may bias evaluation toward ``detectable'' cases. Finally, curriculum learning uses supervised error during training, which may limit applicability in settings without reliable ground-truth gaze.

At the same time, FIUD does not require quality factors at inference and thus avoids deployment-time dependence on landmark detectors. In settings where supervised error is unavailable, the distillation and ranking components remain applicable whenever a proxy teacher signal is available.

\section{Conclusions}
We presented Factor-Informed Uncertainty Distillation (FIUD), a teacher-student approach that leverages interpretable image-quality cues to improve reliability of gaze estimation. FIUD distills factor-based error structure into a single-pass uncertainty head, improving uncertainty ranking and selective prediction utility, particularly in unconstrained data. By shaping uncertainty to align with actionable capture conditions while retaining single-pass efficiency, FIUD supports practical system behaviors such as abstention, user feedback, and robust gaze-aware interaction.

This factor-informed approach suggests a few extensions worth pursuing. The gradient-boosting teacher could be swapped for a model that adapts online, drawing on sensor-driven recognition methods~\cite{andreu2010real, andreu2013evolving}, so that it does not need retraining offline every time deployment conditions change. The quality factors used here are also computed per frame; a temporal reasoning layer~\cite{kiani2022temporal} could track error across a gaze sequence instead, and eye-movement feature-extraction tooling~\cite{andreu2016ealab} could widen the factor vocabulary beyond illumination, sharpness, and eye visibility. It may also help to express quality factors as linguistic variables rather than numeric ones, following computing-with-words methodologies~\cite{gupta2022gentle}, or to build the teacher itself on explainable fuzzy representation learning~\cite{fumanal2023artxai}, so the quality-to-error mapping stays easy for practitioners to audit. None of this is specific to gaze: pairing an interpretable teacher with a lightweight single-pass student is a pattern that fits any reliability-critical setting where failure modes are easy to name but costly to compute, including clinical brain-computer interfaces~\cite{chowdhury2021clinical} and deep learning systems generally that need to stay reliable on noisy, real-world input~\cite{chaudhary2023review}.

\begin{acks}
This research is kindly supported by UKRI BBSRC project EyeWarn (code: APP37953). We also thank the ETRA '26 reviewers for their constructive feedback, which helped improve the clarity and presentation of this work. Finally, we are grateful to our colleagues for helpful discussions and support throughout the project.
\end{acks}

\bibliographystyle{ACM-Reference-Format}
\bibliography{references}

\end{document}